\documentclass{article}

\usepackage[final,nonatbib]{score_workshop}
\usepackage[numbers]{natbib}
\usepackage[utf8]{inputenc} %
\usepackage[T1]{fontenc}    %
\usepackage{hyperref}       %
\usepackage{url}            %
\usepackage{booktabs}       %
\usepackage{amsfonts}       %
\usepackage{nicefrac}       %
\usepackage{microtype}      %
\usepackage{xcolor}         %
\usepackage{caption}
\usepackage{subcaption}

\usepackage{mathtools}
\usepackage{amsthm}
\usepackage{algorithm}
\usepackage{algpseudocode}
\usepackage{amsmath}
\usepackage{bbm}
\usepackage{amssymb}
 \usepackage{amsfonts,mathrsfs,dsfont}

\newcommand{\emb}{\mathrm{emb}}
\newcommand{\enc}{\mathrm{enc}}
\newcommand{\tf}{t_{\mathrm{fixed}}}
\theoremstyle{plain}

\theoremstyle{definition}

\theoremstyle{remark}

\newtheoremstyle{named}{}{}{\itshape}{}{\bfseries}{.}{.5em}{Theorem #3}
\theoremstyle{named}

\title{Multiresolution Textual Inversion}

\author{%
  Giannis Daras \\
  Department of Computer Science\\
  UT Austin\\
  \texttt{giannisdaras@utexas.edu} \\
  \And
  Alexandros G. Dimakis \\
  Department of Electrical and Computer Engineering \\
  UT Austin \\
  \texttt{dimakis@austin.utexas.edu}
}

\begin{document}

\maketitle

\begin{abstract}
We extend Textual Inversion to learn pseudo-words that represent a concept at different resolutions. This allows us to generate images that use the concept with different levels of detail and also to manipulate different resolutions using language.
Once learned, the user can generate images at different levels of agreement to the original concept; ``A photo of $S^*(0)$'' produces the exact object while the prompt ``A photo of $S^*(0.8)$'' only matches the rough outlines and colors. 
Our framework allows us to generate images that use different resolutions of an image (e.g. details, textures, styles)  as separate pseudo-words that can be composed in various  ways. 
We open-soure our code in the following URL: \href{https://github.com/giannisdaras/multires_textual_inversion}{https://github.com/giannisdaras/multires\_textual\_inversion}.
\end{abstract}
\vspace{-0.4cm}

\section{Introduction}

Textual Inversion~\citep{textual_inversion} is a novel technique for introducing a new concept in a pre-trained text conditional generative model. Given a few images of a user-provided concept (e.g. an object or style), Textual inversion learns a new ``word" in the embedding space to represent that object. Remarkably, these new \textit{pseudo-words} can be composed in language to produce all kinds of creative compositions. 

\begin{figure}[ht]
    \centering
    \subcaptionbox*{Inputs}[0.23\textwidth]{}
    \subcaptionbox*{\texttt{A painting of a dog in the style of <jane({$\tf$})>.}}[0.69\textwidth]{}
    \subcaptionbox*{}[0.23\textwidth]{\includegraphics[width=0.23\textwidth]{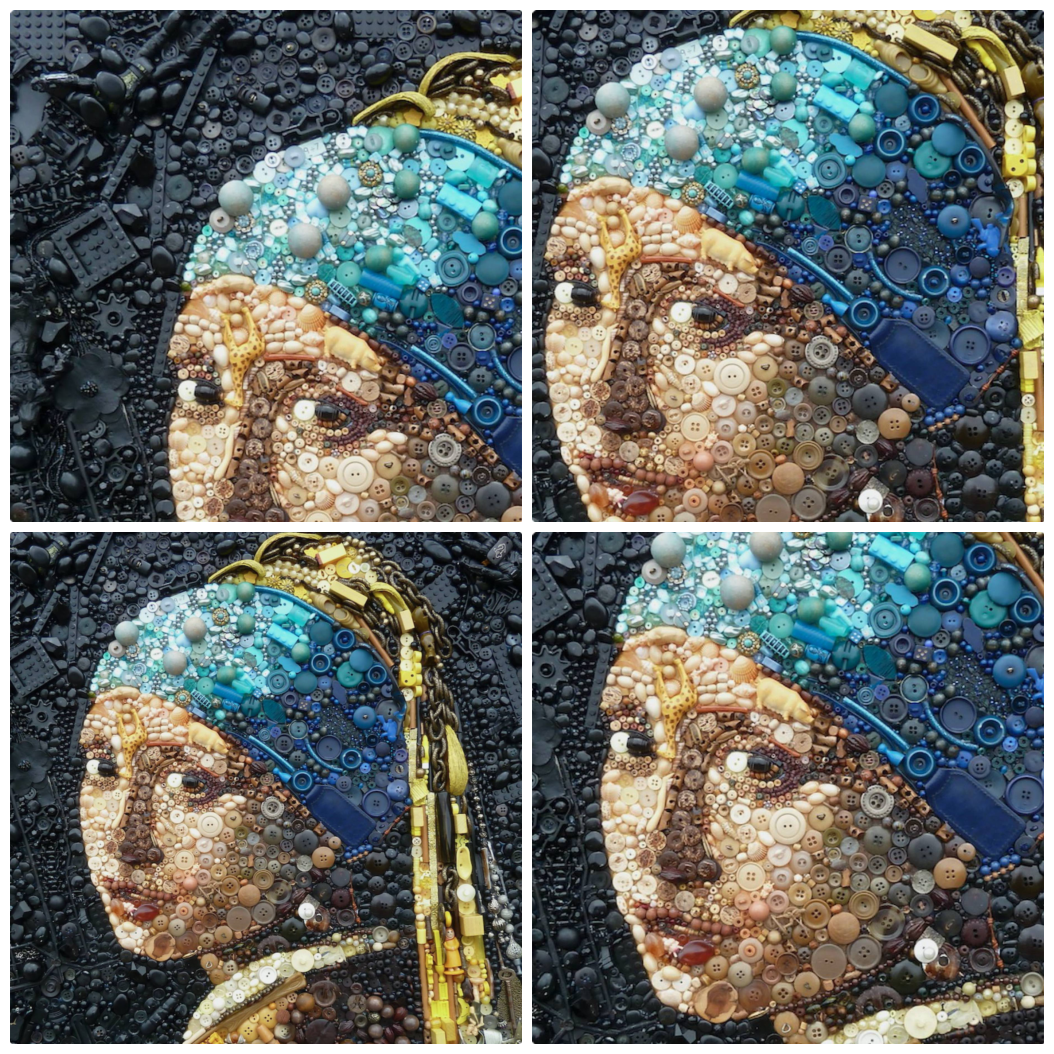}}
    \subcaptionbox{$\tf=0.0$}[0.23\textwidth]{\includegraphics[width=0.23\textwidth]{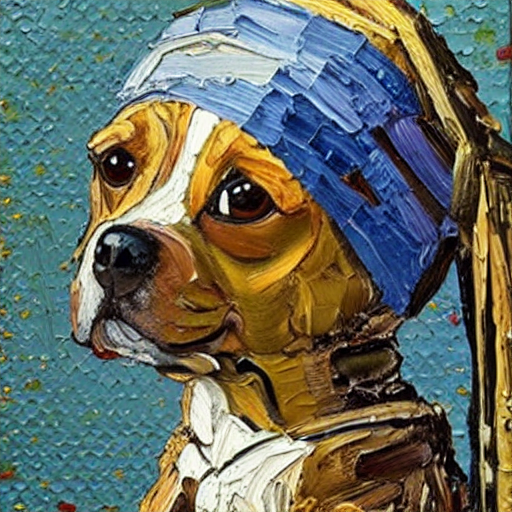}}
    \subcaptionbox{$\tf=0.5$}[0.23\textwidth]{\includegraphics[width=0.23\textwidth]{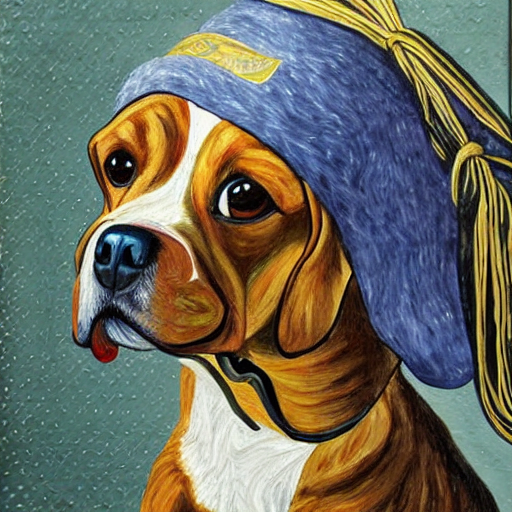}}
    \subcaptionbox{$\tf=0.7$}[0.23\textwidth]{\includegraphics[width=0.23\textwidth]{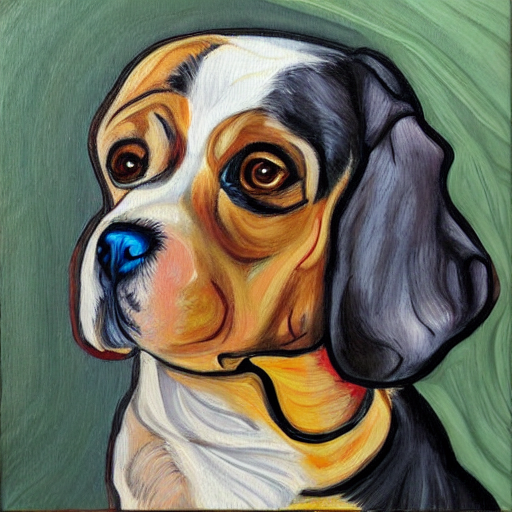}}
    \caption{\small{Multi-resolution textual inversion. We learn with our method the input concept as \texttt{<jane>}. We then show images generated with the prompt: \texttt{A painting of a dog in the style of <jane({$\tf$})>}, where $\tf$ controls the resolution in which the pseudo-word captures the input concept. Each pseudo-word represents the concept with different levels of detail with the zero index capturing the full detailed learned concept.}}
    \label{fig:figure1}
\end{figure}

We extend Textual inversion to learn multiple pseudo-words that represent a concept at different resolutions. 
For example, in Fig. \ref{fig:figure1} the input concept is the 
re-creation, with various small objects, of Vermeer's \textit{Girl with a Pearl Earring}, by artist J. Perkins~\citep{janeperkins}.
We learn this concept using 4 re-croppings of the input.
We then use a pre-trained text-to-image model to generate images using prompts that contain the learned pseudo-words that represent the concept at various resolutions. 

\section{Method}
\paragraph{Background.}
Text-conditional diffusion models are trained to restore a clean image, $x$, based on a corrupted observation, $z_t(x)$, and a text caption, $y$, describing the image~\citep{dalle2, imagen, ldms, ho2020denoising, song2020scorebased}. The corruption, which is typically additive noise, can be either in the image itself~\citep{dalle2, imagen} or in an encoding of the image~\citep{ldms}. We denote with $z_t(x)$ the diffused observation of image $x$ at time $t$. Text-conditional models are typically trained to minimize the objective:
\begin{gather}
    J(\theta) = \mathbb E_{t\sim \mathcal U[0,1]} \mathbb E_{(x, y) \sim p}\mathbb E_{z_t\sim q_t(z_t|x, t)}||\hat \epsilon_{\theta}(z_t(x), t, c_{\theta}(y)) - \epsilon(z_0, z_t)||^2,
\end{gather}
where $p$ is the joint distribution of image-captions, $q_t$ is the distribution of the diffused observations, $y$ is the text-conditioning and $\epsilon$ is the unscaled residual to $z_0(x)$.

The conditioning network, $c_{\theta}(y)$, maps the tokenized text $y$ into a latent that is used to inform the prediction of the denoiser network, $\hat \epsilon_{\theta}$. The first layer of $c_{\theta}(y)$ is typically a lookup table, $\mathrm{emb}(y)$, that maps tokens to vectors. We decompose the conditioning network as: $c_{\theta}(y) = \mathrm{enc}_{\theta}(\mathrm{emb}_{\theta}(y))$.

The authors of Textual Inversion~\citep{textual_inversion} consider the task of finding an embedding, $\mathrm{emb}^*$ for a pseudo-token $y^*$ that represents a concept described by a small collection of images $X = \{x_1, ..., x_N\}$. They do so by solving the following optimization problem:

\begin{gather}
    \min_{\emb}\mathbb E_{t\in \mathcal U[0, 1]}\mathbb E_{x \sim \mathcal U_X} \mathbb E_{z_t\sim q_t(z_t|x, t)}\left|\left| \hat \epsilon_{\theta}(z_t(x), t, \enc_{\theta}(\emb)) - \epsilon(z_0, z_t)\right|\right|^2.
    \label{eq:textual_obj}
\end{gather}
Intuitively, Textual Inversion is optimizing for an embedding that helps the model predict in the most effective way the residual across all the images in the set and across all corruption levels.

\paragraph{Multiresolution Textual Inversion.} 
The key idea of our method is that \textit{the conditioning signal can depend on the diffusion time (i.e. the noise level)}. For example, if the input image to the diffusion model is a slightly noisy image of a cat, the text conditioning should give information on the details of the cat (e.g. texture) and not necessarily on the image class which can easily be inferred from the image input. Similarly, for very noisy images the conditioning should be capturing basic information such as image class (e.g. cat vs dog) and colors. In general, we propose the following objective:

\begin{gather}
    \min_{\{\emb_0, ..., \emb_{T-1}\}}\mathbb E_{t\sim \mathcal U[0, 1]}\mathbb E_{x \sim \mathcal U_X} \mathbb E_{z_t\sim q_t(z_t|x, t)}\left|\left| \hat \epsilon_{\theta}(z_t(x), t, \enc_{\theta}(\emb_{\lfloor t/T \rfloor})) - \epsilon(z_0, z_t)\right|\right|^2.
    \label{eq:ours}
\end{gather}

After training, we have learned a set of embeddings, $\mathrm{Emb}^*=\{\emb_{0}, ..., \emb_{T-1}\}$. Each of the the elements of the set, captures different levels of detail. 
The idea of having latents that describe the image with different amounts of agreement to the input has been exploited to solve inverse problems with GANs~\citep{ilo, sgilo, gan_surgery}. In these works, the resolution is controlled by the index of the GAN layer in which we invert our images. In our method the agreement to the input is determined by the diffusion time index and we can manipulate the different resolutions using language.

We note that our method can also be extended to support DreamBooth~\citep{ruiz2022dreambooth}, an alternative method for Textual Inversion that finetunes the whole model.

We can create embeddings indexed by continuous time by using linear interpolations of the elements of the learned set.
We present three different ways to use the learned embeddings to sample at different levels of agreement to the concept appearing in the input images. For a visual comparison, we show how these methods perform for a given image at different resolution levels, in Fig. \ref{fig:sampling_comparisons}.

\paragraph{Fixed Resolution Sampling (Method 1).} The first method fixes the conditioning to a specific embedding throughout the sampling. It is the most similar sampling method to Textual Inversion, in the sense that it is using a fixed embedding for all diffusion levels. The user can control the resolution by picking one embedding from the set $\mathrm{Emb}^*$. This method allows the user to visualize what is learned at each resolution. As we explained earlier, we expect that embeddings that correspond to time close to $t=0$, should learn details (e.g. texture) since those are more informative to denoise a slightly noisy image than image class for example. Embeddings closer to $t=1$ should be related to more coarse information about the image, e.g. what is the object, what are the colors, etc.

For the next two sampling methods, we change the conditioning based on the sampling time. Assume that we want to generate an image at resolution $t=t_{\mathrm{fixed}}$. We propose two sampling methods:

\paragraph{Semi Resolution-Dependent Sampling (Method 2).} Semi Resolution-Dependent Sampling uses $\emb_{t}$ when the sampling time is $t$ if $t \geq t_{\mathrm{fixed}}$ and no conditioning otherwise. Essentially, this forces the model to generate images that match (in a distributional sense) the input images at noise level $t$. This method is particularly useful for style-transfer and creative prompts -- for some percentage of the sampling procedure the model performs unconditional generation (with a starting point a diffused sample that matches the original concept with noise). This idea extends the novel SDEdit paper~\citep{sdedit} in the sense that it allows to do guided image synthesis (but this time starting from language tokens).

\paragraph{Fully Resolution-Dependent Sampling (Method 3).} Fully Resolution-Dependent Sampling is similar to Semi Resolution-Dependent Sampling, but instead of doing unconditional generation for $t<t_{\mathrm{fixed}}$, it fixes the embedding to $\emb_{\mathrm{fixed}}$. This still allows for variations from the given concept (since $\emb_{\mathrm{fixed}}$ only resembles it in a given resolution), but the variations are more controlled since there is no sampling period of unconditional generation.

\paragraph{Creating a textual interface.} Similar to Textual Inversion, we create pseudo-words that allow us to use language to guide the image generation process with our learned embeddings. The difference is that the embeddings for our pseudo-words might be changing based on the sampling time. We use $S^*|\tf|, S^*(\tf), S^*[\tf]$ to denote tokens that are used with Fixed Resolution, Semi Resolution and Fully Resolution Dependent Sampling respectively.

\section{Experiments}
\begin{figure}[!htp]
    \centering
    \captionsetup[subfigure]{aboveskip=-7pt,belowskip=-1pt}
    \subcaptionbox*{\label{fig:button_inputs}}{\includegraphics[width=0.18\textwidth]{figures/jane/jane.png}}
    \subcaptionbox*{\label{fig:cat_inputs}}{\includegraphics[width=0.18\textwidth]{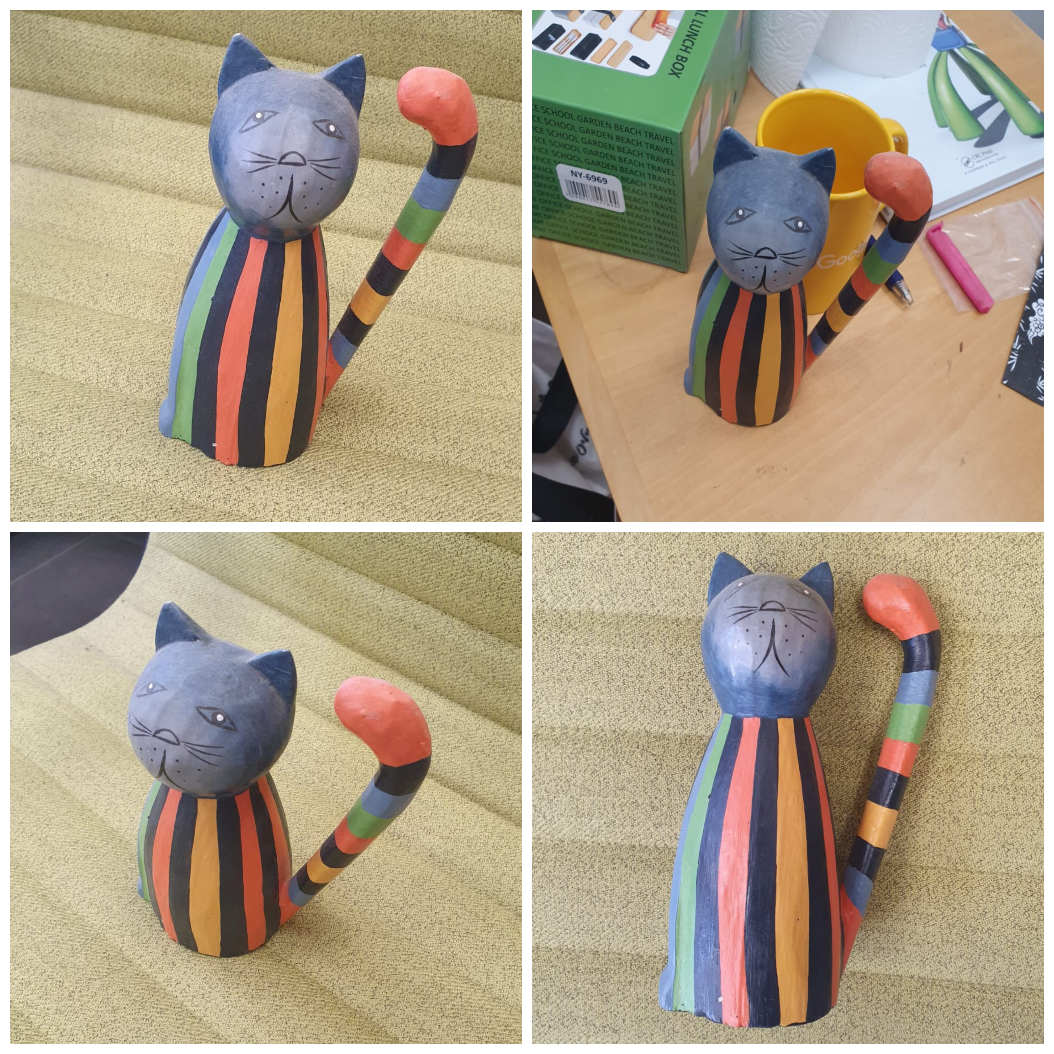}}
    \subcaptionbox*{}{\includegraphics[width=0.18\textwidth]{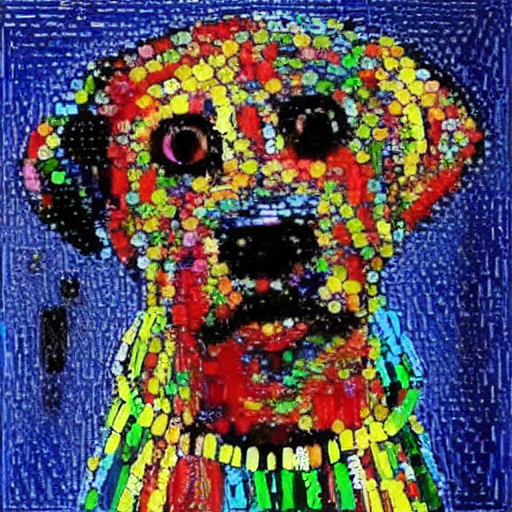}}
    \subcaptionbox*{}{\includegraphics[width=0.18\textwidth]{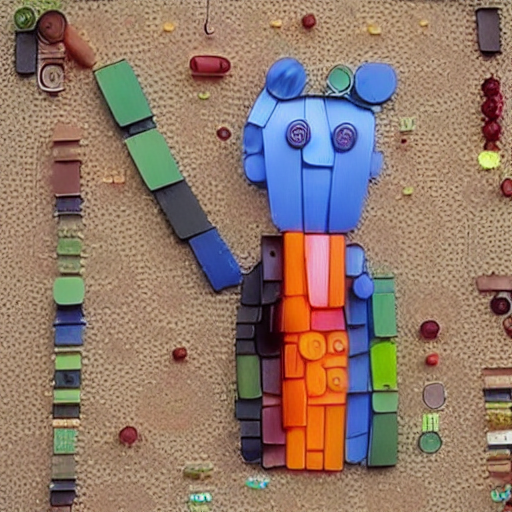}}
    \subcaptionbox*{}{\includegraphics[width=0.18\textwidth]{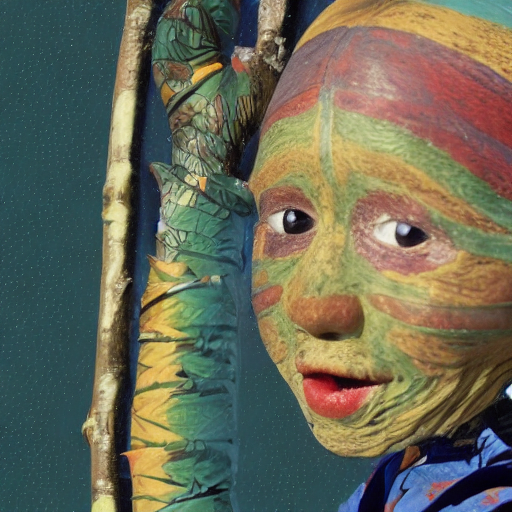}}
    \subcaptionbox{Inputs learned as <jane> and <cat>.\label{fig:button_cat_inputs}}[0.36\textwidth]{}
    \subcaptionbox{\tiny{A painting of a dog made with <jane$|0.1|$>.}}[0.18\textwidth]{}
    \subcaptionbox{\tiny{A photo of <cat$(5)$> made with <jane$|0.1|$>.}}[0.18\textwidth]{}
    \subcaptionbox{\tiny{A photo of <jane$(3)$> similar to <cat$[0]$>.}}[0.18\textwidth]{}
    \caption{\small{Compositions of learned pseudo-words across resolutions.
    Observe that the dog (b) and cat (c) are made from small plastic objects, i.e. using the detailed structure of <jane>, but in the shape of a dog or the toy cat. In (d) the color stripes are obtained from the cat detail.
    }}
    \label{fig:mixes}
\end{figure}

\begin{figure}[ht]
    \centering
    \captionsetup[subfigure]{aboveskip=-8pt,belowskip=-1pt}
    \subcaptionbox*{Inputs}[0.15\textwidth]{}
    \subcaptionbox*{\tiny{\texttt{A photo of $S^*|{\tf}|$} (Fixed Resolution Sampling)}}[0.45\textwidth]{}
    
    \subcaptionbox*{}{\includegraphics[width=0.15\textwidth]{figures/jane/jane.png}}
    \subcaptionbox*{}{\includegraphics[width=0.15\textwidth]{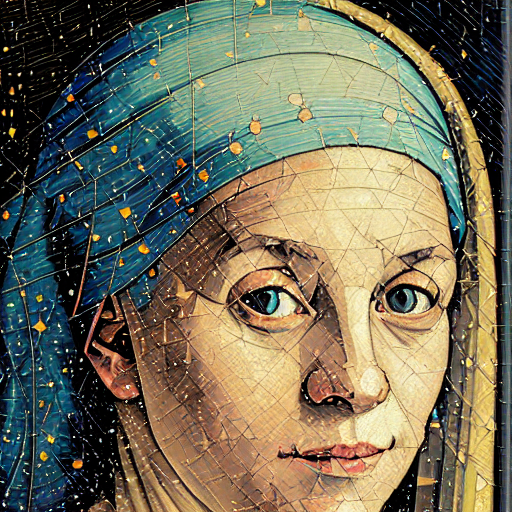}}
    \subcaptionbox*{}{\includegraphics[width=0.15\textwidth]{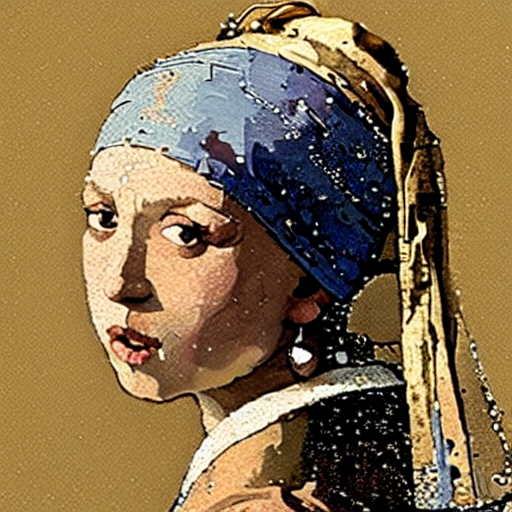}}
    \subcaptionbox*{}{\includegraphics[width=0.15\textwidth]{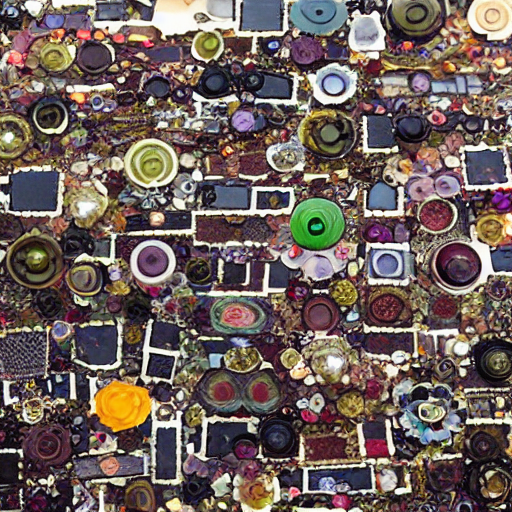}}
    
    \subcaptionbox*{}[0.15\textwidth]{}
    \subcaptionbox*{\tiny{\texttt{A photo of $S^*({\tf})$} (Semi Resolution-Dependent Sampling})}[0.45\textwidth]{}
    
    \subcaptionbox*{}[0.15\textwidth]{}
    \subcaptionbox*{}{\includegraphics[width=0.15\textwidth]{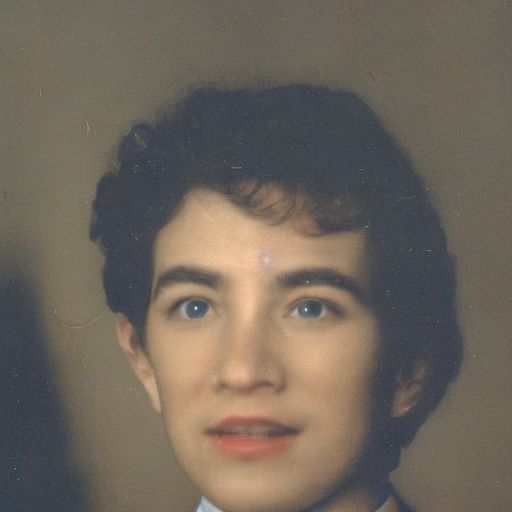}}
    \subcaptionbox*{}{\includegraphics[width=0.15\textwidth]{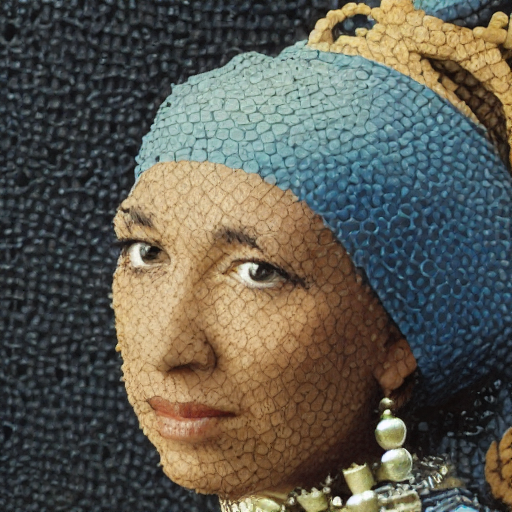}}
    \subcaptionbox*{}{\includegraphics[width=0.15\textwidth]{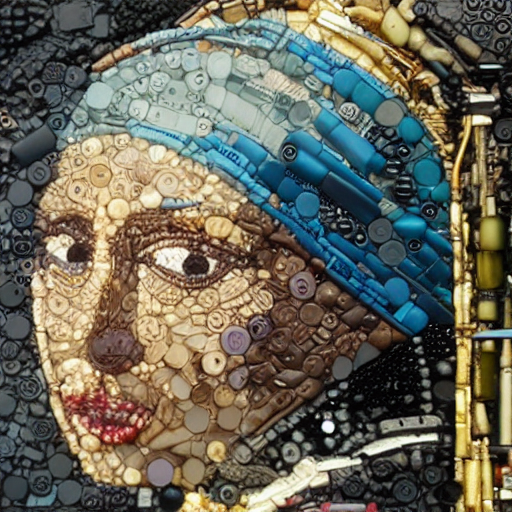}}
    
    \subcaptionbox*{}[0.15\textwidth]{}
    \subcaptionbox*{\tiny{\texttt{A photo of $S^*[{\tf}]$} (Fully Resolution-Dependent Sampling)}}[0.45\textwidth]{}
    \captionsetup[subfigure]{aboveskip=5pt,belowskip=0pt}

    \subcaptionbox*{}[0.15\textwidth]{}
    \subcaptionbox{$\tf=0.8$}{\includegraphics[width=0.15\textwidth]{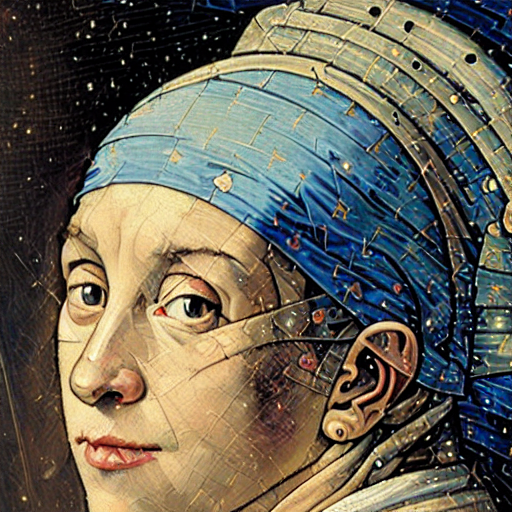}}
    \subcaptionbox{$\tf=0.5$}{\includegraphics[width=0.15\textwidth]{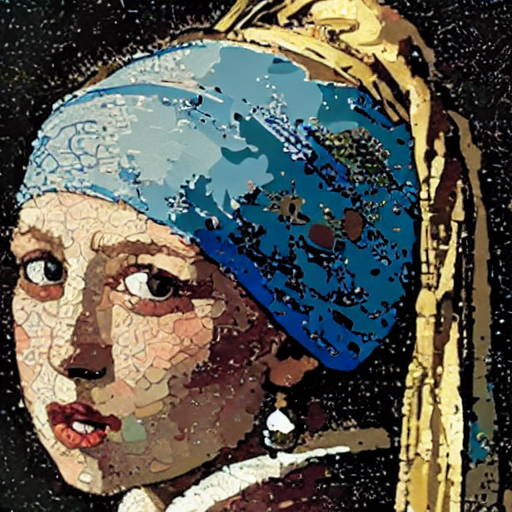}}
    \subcaptionbox{$\tf=0.0$}{\includegraphics[width=0.15\textwidth]{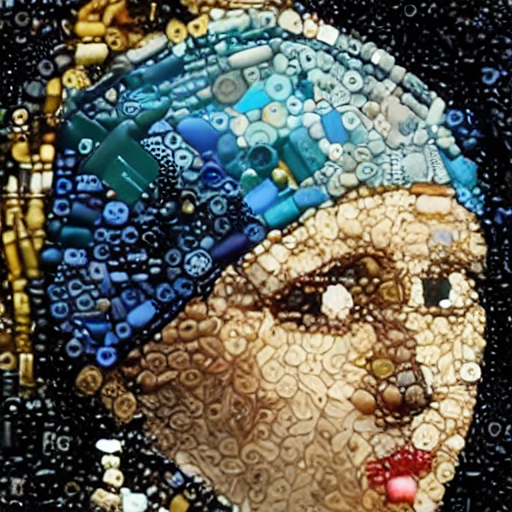}}
    \caption{\small{Comparison of different sampling methods that be used for Multiresolution Textual Inversion. Each row shows a different sampling algorithm and each column a different resolution. With Fixed Resolution Sampling (row 1), we can selectively extract details of the image, e.g. $S^*|0.0|$ extracts only the buttons. Semi Resolution-Dependent Sampling (row 2) allows for larger variations from the input for high values of $\tf$. Fully Resolution-Dependent Sampling (row 3) maintains higher agreement to the inputs along all resolutions.}}
    \label{fig:sampling_comparisons}
\end{figure}

We begin by comparing the different sampling methods. Results are shown in Fig. \ref{fig:sampling_comparisons}. Each row shows a different sampling algorithm each each column a different resolution. With Fixed Resolution Sampling (row 1), we can selectively extract details of the image, e.g. $S^*_{0.0}$ extracts only the buttons. Semi Resolution-Dependent Sampling (row 2) allows for larger variations from the input for high values of $\tf$, e.g. $S^*(0.8)$ gives a photorealistic image of person that only roughly preserves colors and pose from the input concept. Finally, Fully Resolution-Dependent Sampling (row 3) maintains higher agreement to the input concept along all resolutions.

\begin{figure}[ht]
    \begin{subfigure}{0.30\textwidth}
        \caption*{\large{Inputs}}
    \end{subfigure}
    \begin{subfigure}{0.3\textwidth}
        \caption*{\large{Ours}}
    \end{subfigure}
    \begin{subfigure}{0.3\textwidth}
        \caption*{\large{Textual Inversion}}
    \end{subfigure}
    \subcaptionbox{}{\includegraphics[width=.3\textwidth]{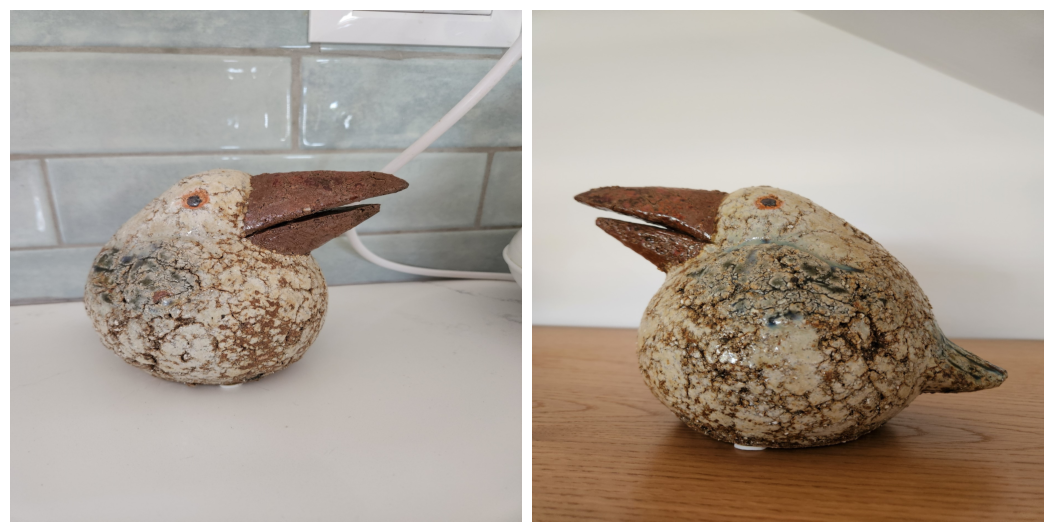}}
    \subcaptionbox{\scriptsize{Watercolor painting of $S^*$ on a branch.}}{\includegraphics[width=.15\textwidth]{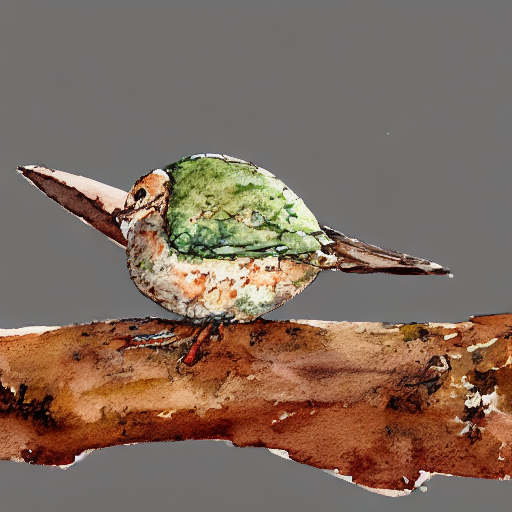}}
    \subcaptionbox{\scriptsize{Grainy photo of $S^*$ in angry birds.}}{\includegraphics[width=.15\textwidth]{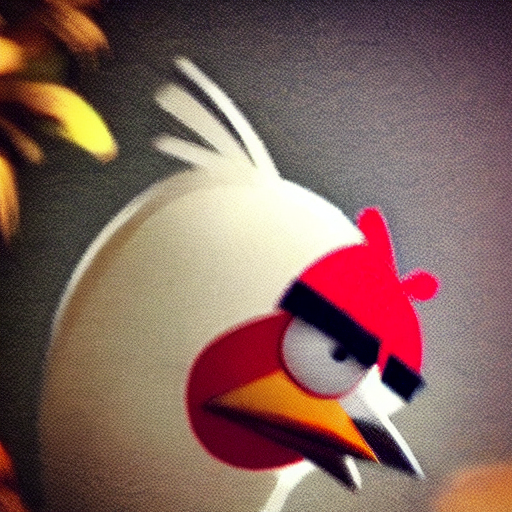}}
    \subcaptionbox{\scriptsize{Watercolor painting of $S^*$ on a branch.}}{\includegraphics[width=.15\textwidth]{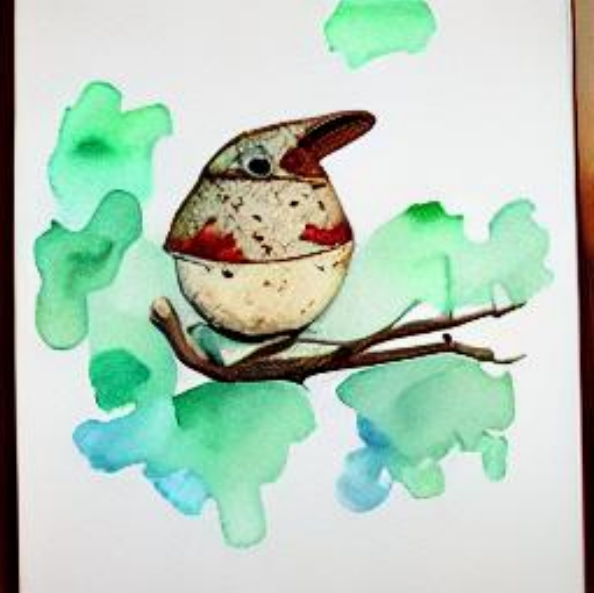}}
    \subcaptionbox{\scriptsize{Grainy photo of $S^*$ in angry birds.}}{\includegraphics[width=.15\textwidth]{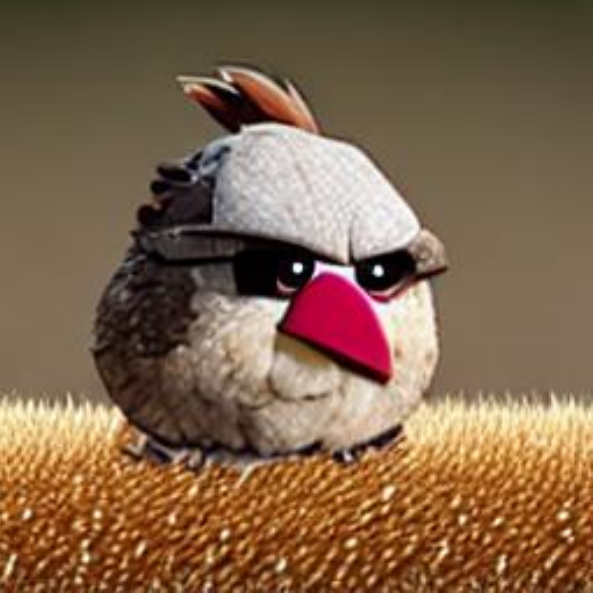}}

    \subcaptionbox{}{\includegraphics[width=.3\textwidth]{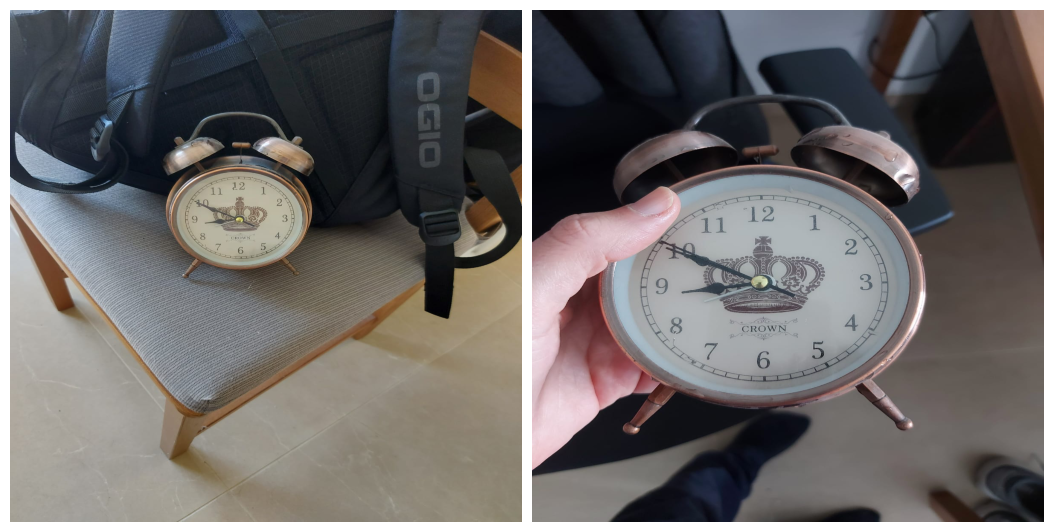}}
    \subcaptionbox{\scriptsize{A photo of $S^*$ on the beach.}}{\includegraphics[width=.15\textwidth]{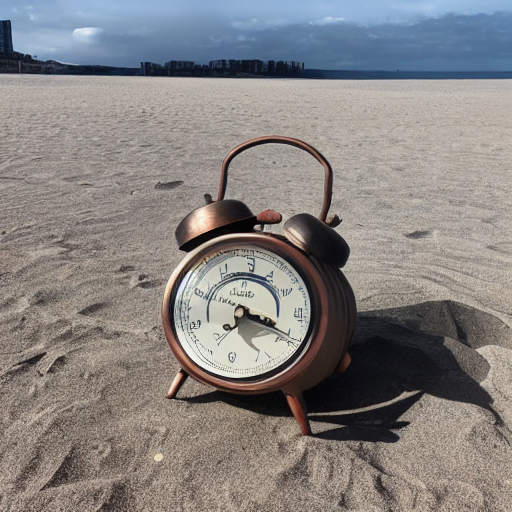}}
    \subcaptionbox{\scriptsize{A photo of $S^*$ on the moon.}}{\includegraphics[width=.15\textwidth]{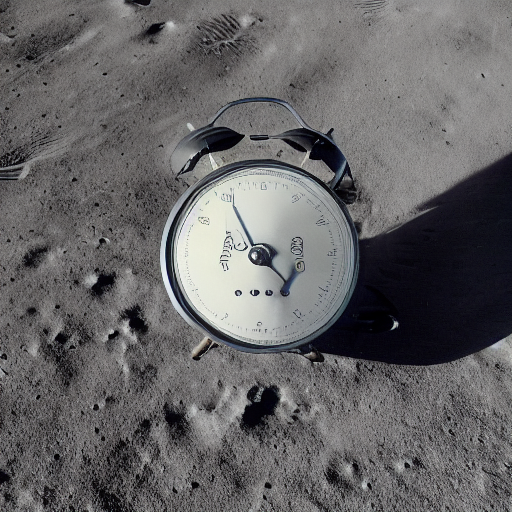}}
    \subcaptionbox{\scriptsize{A photo of $S^*$ on the beach.}}{\includegraphics[width=.15\textwidth]{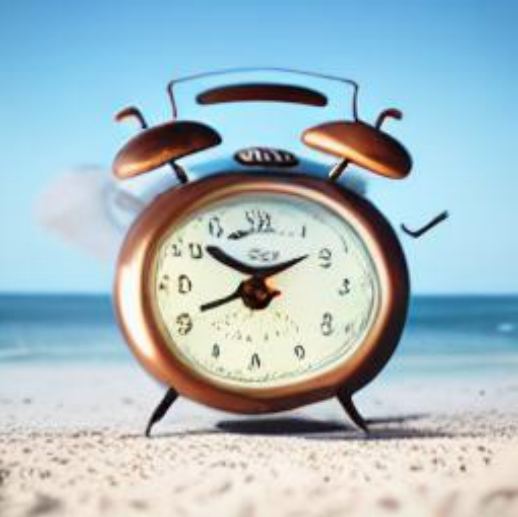}}
    \subcaptionbox{\scriptsize{A photo of $S^*$ on the moon.}}{\includegraphics[width=.15\textwidth]{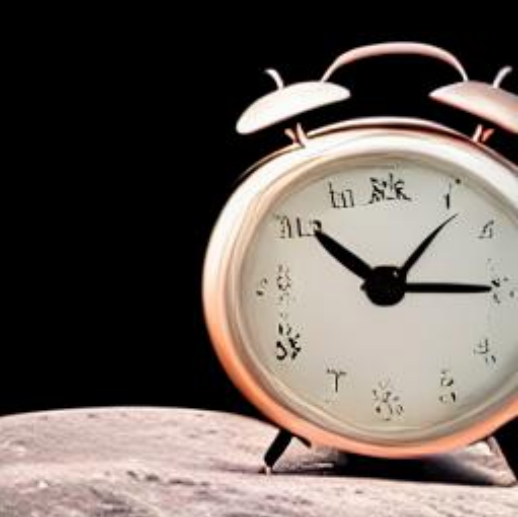}}

    \subcaptionbox{}{\includegraphics[width=.3\textwidth]{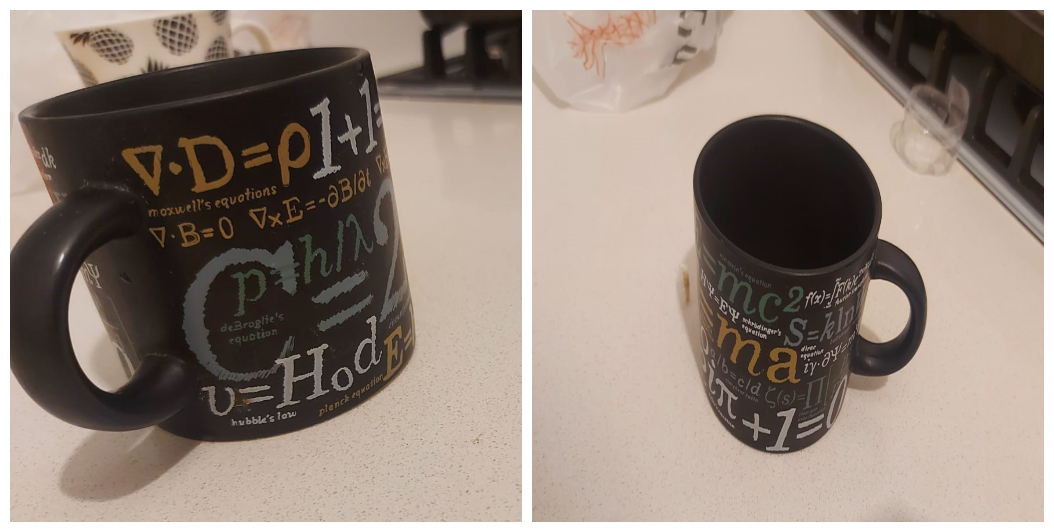}}
    \subcaptionbox{\scriptsize{A photo of $S^*$ on the beach.}}{\includegraphics[width=.15\textwidth]{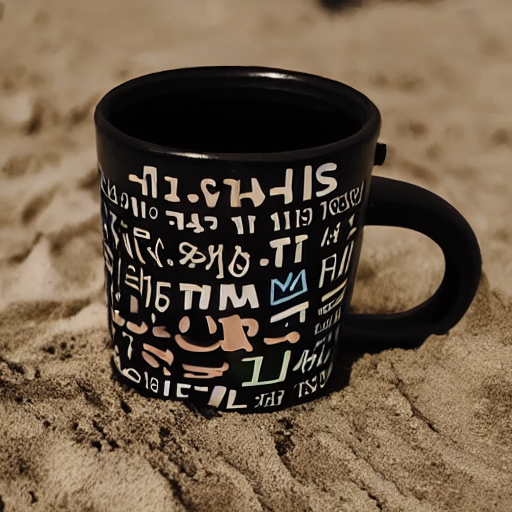}}
    \subcaptionbox{\scriptsize{A photo of $S^*$ on the moon.}}{\includegraphics[width=.15\textwidth]{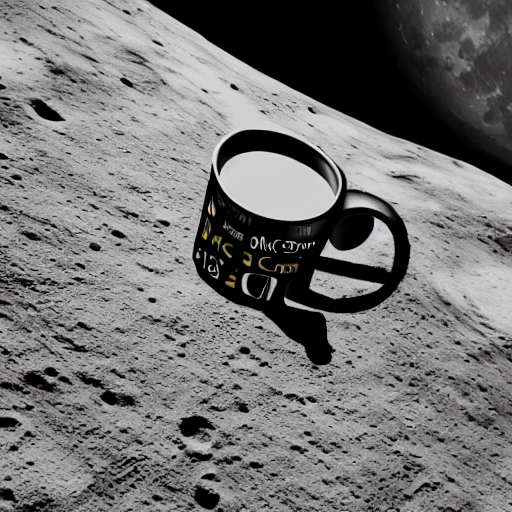}}
    \subcaptionbox{\scriptsize{A photo of $S^*$ on the beach.}}{\includegraphics[width=.15\textwidth]{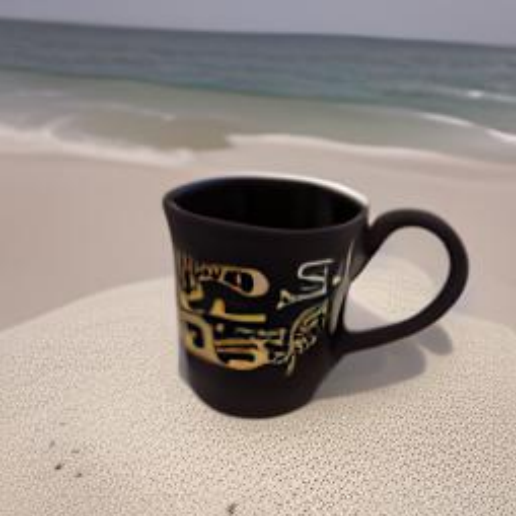}}
    \subcaptionbox{\scriptsize{A photo of $S^*$ on the moon.}}{\includegraphics[width=.15\textwidth]{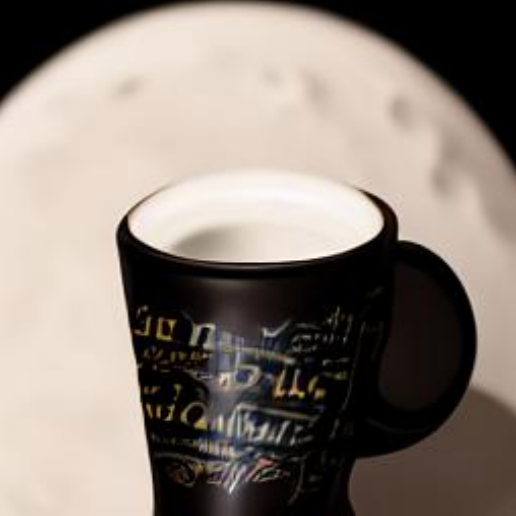}}

    \caption{\small{Personalized image generation. We show that with our method we can use the pseudo-words for the learned concept to create personalized images as if it was a normal word token. Our method performs on par and sometimes better than Textual Inversion on this task.}}
    \label{fig:compositions}
\end{figure}

We then show that Multiresolution Textual Inversion performs on par (or even better) with Textual Inversion. Fig. \ref{fig:compositions} shows that our method can use the learned concepts in combinations with prompts of all sorts -- we use the prompt and images from the Textual Inversion paper.

Finally, Fig. \ref{fig:mixes} shows that the learned pseudo-words can be combined in arbitrary ways across different resolutions and concepts. For example, we can extract the plastic object detailed structure of concept <jane> and use it to generate other objects, such as a dog painting and the cat of Fig. \ref{fig:button_cat_inputs}.

\section{Conclusions and Future Work}
We showed how to learn multiple pseudo-words representing an input concept at different scales. This expands the prompt vocabulary and can be easily used with pre-trained text conditional diffusion models. 
As future work, we would like to quantitatively measure if our method outperforms Textual Inversion and if models trained with diffusion processes beyond additive noise \citep{bansal2022cold, soft_diffusion} can
extract different aspects of scale, depending on how they corrupt their inputs.

\clearpage
\begin{ack}
This research has been supported by NSF Grants CCF 1763702,
AF 1901292, CNS 2148141, Tripods CCF 1934932, IFML CCF 2019844, the Texas Advanced Computing Center (TACC) and research gifts by Western Digital, WNCG IAP, UT Austin Machine Learning Lab (MLL), Cisco, Onassis Fellowship, Bodossaki Fellowship and the Archie Straiton Endowed Faculty Fellowship.
\end{ack}

\bibliography{references.bib}
\bibliographystyle{unsrtnat}

\appendix

\end{document}